\documentclass[letterpaper, 10 pt, conference]{ieeeconf}  

\IEEEoverridecommandlockouts                              
\usepackage{booktabs}
\usepackage{tabularx}
\usepackage{makecell}

\title{\LARGE \bf
audEERING's approach to the One-Minute-Gradual Emotion Challenge
}

\author{Andreas Triantafyllopoulos, Hesam Sagha, Florian Eyben, Bj\"orn Schuller \\
audEERING GmbH, Gilching, Germany
}
\begin{document}
\maketitle
\thispagestyle{empty}
\pagestyle{empty}

\begin{abstract}
This paper describes audEERING's submissions as well as additional evaluations for the One-Minute-Gradual (OMG) emotion recognition challenge.
We provide the results for audio and video processing on subject (in)dependent evaluations. On the provided Development set, we achieved 0.343 Concordance Correlation Coefficient (CCC) for arousal (from audio) and .401 for valence (from video).
\end{abstract}
\section{INTRODUCTION}
The OMG dataset consists of 5288 (train: 2442, dev: 617, test: 2229) segments from YouTube videos of about 1-minute each, and the raters annotated some segments in each video on arousal (activation) [0..1] and valence [-1..1] dimensions. For the sake of consistency we mapped arousal also to [-1..1] range.

In our approach, we use cross-fold validation on the training data, train different models which are optimized on out-of-fold predictions, and finally use these models on the official development set and take average their outputs. However, since in the challenge there was significant video- (and speaker-) overlap between the train, development, and test partitions, the results of each fold may be biased toward specific speakers. Therefore, we performed two sets of analyses: i) random shuffling of the samples in the training set, and ii) arranging speaker independent folds, in which we manually sorted different speakers into five folds, and made each of them speaker independent. In the later case, data were sorted into 74 disjoint speakers. Moreover, on our analyses, we always scale the output to have the same mean and variance as the training folds; in this way the CCC could increase up to Correlation Coefficient (CC).

Regarding the available modalities, we only used audio for arousal and valence, and video for valence separately, and in one submission we combined the results of these two modalities for valence. In the following we explain our processing for audio and video and their corresponding results.
\section{AUDIO PROCESSING}
We used the openSMILE toolkit \cite{eyben2013recent}\footnote{https://audeering.com/technology/opensmile/} to extract 1170 features (a reduced version of the ComParE 2016 feature set \cite{weninger2013acoustics}) in which it consists of 9 Functionals (e.g., mean, variance) over 65 Low-Level-Descriptors (e.g., spectrogram, pitch) which were extracted with the window size of 0.06s and the step of 0.01s. The Functionals were computed from LLDs over a window of 2s with the step of 1s. The set of LLDs and Functionals are provided in Table~\ref{tab:lld_func}.
\begin{table}[t]
\caption{Set of Low-Level Descriptors (LLDs) and Functionals and their groups.}
\begin{tabular*}{\linewidth}{l@{\extracolsep{\fill}}c}\toprule
\multicolumn{2}{c}{ENERGY RELATED LLDs}\\ \midrule
Sum of auditory spectrum (loudness)&	Prosodic\\
Sum of RASTA-style filtered auditory spectrum&	Prosodic\\
RMS energy, zero-crossing rate&	Prosodic\\\toprule
\multicolumn{2}{c}{55 SPECTRAL LLD}\\ \midrule
RASTA-style auditory spectrum, bands 1--26 (0--8 kHz)	&Spectral\\
MFCC 1--14	&Cepstral\\
Spectral energy 250--650 Hz, 1--4 kHz&	Spectral\\
Spectral roll off point 0.25, 0.50, 0.75, 0.90&	Spectral\\
Spectral flux, centroid, entropy, slope&	Spectral\\
Psychoacoustic sharpness, harmonicity&	Spectral\\
Spectral variance, skewness, kurtosis&	Spectral\\\toprule
\multicolumn{2}{c}{6 VOICING RELATED LLD}\\\midrule
F$_0$ (SHS and viterbi smoothing)&	Prosodic\\
Prob. of voice&	Sound quality\\
Log. HNR, Jitter (local, delta), Shimmer (local)	&Sound quality\\
\toprule
\multicolumn{2}{c}{FUNCTIONALS APPLIED TO LLD/$\Delta$LLD}\\\midrule
Arithmetic mean	&Moments\\
Standard deviation	&Moments\\
Linear regression slope,  quadratic error&	Regression\\
Quadratic regression a, quadratic error	&Regression\\
Percentile range 1--99\%	&Percentiles\\
6\% Percentile ($\approx$min), 94\% percentile ($\approx$max)&	Percentiles\\
\end{tabular*}
\label{tab:lld_func}
\end{table}

\begin{table*}[!ht]
\centering
\caption{Audio analysis results on the official dev set. ${^\#}$ challenge submission number, ${^*}$ for arousal the fusion is the average of of \#1 and \#2, and for valence it is the fusion of \#3 and Video results.}
\begin{tabular}{l|ccc|ccc}
Approach& \multicolumn{3}{c}{Arousal} & \multicolumn{3}{c}{Valence} \\	\midrule
 & CC & CCC & Scaled CCC &CC & CCC & Scaled CCC \\ \midrule
CDF Adj. + Rand. Shuffle & 0.343 & 0.325 & 0.343$^{\#1}$ & 0.299 & 0.285&0.291\\
CDF Adj. + Rand. Shuffle + Multitasking &0.335&0.331&0.334 &0.312&0.302&0.303$^{\#3}$\\
CDF Adj. + Spk. Independent & 0.299&0.297&0.298$^{\#2}$  &0.233 &0.227&0.232\\
Mean-Var. Norm. + Rand. Shuffle  &0.236&0.215&0.235&0.271&0.251&0.264\\
Fusion${^*}$  &  0.299& 0.298&0.299$^{\#3}$&0.400&0.383&0.388$^{\#2}$\\\midrule
Provided Baseline (Audio w. openSMILE features) &--&0.15&--&--&0.21&--\\
\end{tabular}
\label{tab:aud_res}
\end{table*}

After feature extraction, the Cumulative Distribution Function (CDF) for each partition and each feature were adjusted to have the same CDF as the Normal distribution for each partition separately (train, dev, test). We also tried Mean-Variance Standardization for each partition, however, it suffers from outliers, whereas the features after the CDF adjustment will not have any.

We used six-fold cross validation (randomly shuffled or speaker independent, see introduction for details) to train six models. CURRENNT \cite{weninger2015introducing} was used to train two-layer neural networks with a Bidirectional Long-Short-Term Memory (BLSTM) node of size 100 in the first layer, and an LSTM node of size 40 in the second layer \cite{hochreiter1997long}. The output layer has an identity activation function. The loss function was set to minimize $-CCC$ (similar to \cite{trigeorgis2016adieu}), and the stopping criterion was for the out-of-fold predictions to show no improvement over 15 consecutive epochs. The outputs were scaled to have the same mean and variance as the training folds. Finally, to evaluate the models on the official development set, we averaged their predictions. In the following, we briefly describe the settings for each submission we made for the challenge:\\
{\bf Submission \#1 of arousal}: CDF adjustment, random shuffling of the training data, training the network on only arousal.\\
{\bf Submission \#2 of arousal}: CDF adjustment, speaker independent folding of the training data, training the network on only arousal.\\
{\bf Submission \#3 of arousal}: average of \#1 and \#2.\\
{\bf Submission \#3 of valence}: CDF adjustment, random shuffling of the training data, training the network for both arousal and valence (Multi-tasking).

The results of audio processing (and fusion with video) are summarized in Table~\ref{tab:aud_res}. The best achieved CCC is from the CDF adjustment and random shuffling of the videos.
For the challenge submissions for the official test set, we combined both the training and the development sets and shuffled the videos across seven folds, and the final prediction is the average of the output of the seven trained models.

\section{VIDEO PROCESSING}

The video processing pipeline consists of the following three steps:
\begin{enumerate}
	\item Face detection and alignment using an open source implementation of MTCCN  library \cite{mtccn},
    \item Deep feature extraction using the VGGFace Convolutional Neural Network (CNN) model \cite{vggface},
    \item Utterance based valence prediction using a two-layer BLSTM network.
\end{enumerate}
We will proceed to describe each of the above steps in more detail below.

\subsection{MTCNN Face Detection and Alignment}
A necessary step in our video processing pipeline is accurately extracting and aligning faces for each frame of the video. For this purpose we adopted the widely used Multi-Task Cascaded Convolutional Network approach, for which there are many open-source implementations. The MTCCN library provides state-of-the-art methods for face detection and alignment, and has the added benefit of being very fast and able to identify multiple faces in one image, which allows us to generalize to multi-subject valence prediction. 

In a nutshell, this method utilizes three CNNs. The first one operates on a scale pyramid of the original image and produces a coarse selection of potential targets. These targets are then fed into the second CNN which is responsible for discarding most of the false candidates. Finally, the third CNN makes the final selection and aligns the faces.

\begin{table*}[t]
\centering
\caption{Video analysis results on the official dev set.${^\#}$ challenge submission number}
\begin{tabular}{l|ccc|ccc} Approach& \multicolumn{3}{c}{Arousal} & \multicolumn{3}{c}{Valence} \\ \midrule
5-fold Random Shuffle  & 0.236 & 0.225 & 0.236 & 0.407 & 0.395 & 0.401$^{\#1}$ \\
5-fold Speaker Disjunctive & - & - & - & 0.360 & 0.330 & 0.360 \\
5-fold Speaker Disjunctive \&  Cross-Entropy Maximization & - & - & - & 0.400 & 0.360 & 0.390\\\midrule
Provided Baseline (Vision - Face Channel) &--&0.12&--&--&0.23&--\\
\end{tabular}
\label{tab:vid_res}
\end{table*}

\subsection{VGGFace Features}
Deep features, and in particular CNN intermediate layers, are potential candidates for face image descriptors for emotion recognition tasks, however the OMG Dataset does not contain enough samples for training a deep CNN, so we opted for an open source model that has been trained on a large face dataset. We used the VGGFace network which is based on the VGG-Very-Deep-16 CNN architecture. This model has been trained for face recognition on a large dataset of celebrities. Our intuition is that using one or more intermediate layers of this network would provide abstract representations of the faces in our images that could then be used for valence prediction.

Thus, after using MTCCN for extracting and aligning faces in each video frame, we feed those faces to the VGGFace model and extract our face embeddings. To be consistent with the way the audio modality is processed, we average the output over two-second frames with a one-second overlap. Finally, we experimented with the three last fully connected layers of VGGFace and empirically determined that the second-to-last one, named \textit{fc7}, provided the best prediction results. 

\subsection{BLSTM network for valence prediction}
The OMG-Challenge is formulated as a sequential regression problem. In the last few years, Recurrent Neural Networks, and in particular Long-Short-Term-Memory Networks, have been the method of choice for these kinds of problems. We therefore used an LSTM architecture for predicting valence from the VGGFace features. In our approach, we opted for a Bidirectional LSTM Network, which allows for integrating both future and past dependencies in our predictions. Intuitively, this approach should produce better results since it allows for a facial expression to be viewed in its entirety by our network before making a prediction.

We experimented with a number of BLSTM architectures, consisting of one to four layers. Although deeper networks are usually better than shallow ones, we opted for only a small number of layers because: a) we did not have enough data for training a deep architecture, and b) our face features are already derived from a deep pipeline so they have already benefited from the generalization properties of deep networks. 

Our final submission is based on a two-layer BLSTM network, with the first node being of size 16 and the second one of size 8, followed by a final dense output layer with a \textit{tanh} activation. The network was trained using sequences of face features which correspond to each utterance in our dataset. We then used this network to predict the labels on the validation and test set. This approach does take into account the temporal dependencies of facial expressions in each utterance clip, but fails to account for the fact that each utterance is part of a larger video. We tried to counterbalance this fact by using stateful LSTMs for prediction, thus carrying over the state of our network across each utterance, but the initial results were not significantly better in the validation set, so we did not include them in our submission. In all cases, we trained our networks using $-CCC$ as our loss function, and normalized our features using mean and standard deviation.

\subsection{Experiment Discussion}

During the development stage, we used the official validation set as test set, then created an ensemble of models trained on different 4-fold subsets of a 5-fold split of the training set, using the remaining fold for validation and early stopping. This should make our predictions more robust to overfitting, and additionally provide a measure of how well our networks would perform on the official test set.

We also trained on manually selected speaker disjunctive folds, but found that the performance of our approach was significantly degraded. This could be due to the fact that the VGGFace model has been trained for speaker identification, and could be addressed by either fine-tuning the network for an affective task on a separate dataset, or selecting features from earlier layers. 

We decided to try and counteract this effect by doing a simple form of Domain-Adaptation on our model. Instead of only trying to minimize the $-CCC$, we jointly trained our network for valence prediction and speaker misclassification. This was achieved by augmenting our valence target vector by the appropriate speaker labels, and substituting our final layer with a dense softmax layer of appropriate size. Then, we altered our loss function of $-CCC$ by adding to it the inverse of the categorical cross-entropy over the speaker labels. 

Intuitively, training with this loss function is a simple way to make our network speaker-agnostic. In practice, more sophisticated domain-adaptation methods should be used, since maximizing the categorical cross-entropy can be achieved by a random scrambling of the inputs to the last layer, but our network is shallow enough for the effects to propagate to the two previous layers as well, and thus increase our CCC results to that of the randomly-shuffled folds.

The video analysis results are presented in table~\ref{tab:vid_res}. We achieve a valence value of 0.401 when training on random shuffling of speakers. Our performance reduces to 0.360 when the folds are made speaker disjunctive, but jointly training our network to maximize categorical cross-entropy remedies this effect and increases valence back to 0.390, which is close to what we get when using random folds. Our arousal results were much lower than the ones produced using the audio modality, so we opted to focus on modelling arousal using only audio features instead.

Finally, for our predictions on the official test set, we merged the training and validation sets, and repeated the procedure using 6-fold cross-validation for early stopping and aggregating the results of the resulting six models.

\section{CONCLUSIONS}
It is a well-known fact that audio and speech contribute more in arousal, and facial gestures more in valence. In our analysis, we also reached the same conclusion. We found that adjusting the Cumulative Distribution Function on audio features significantly improves the performance on the official validation set. Moreover, aggregating decisions of models which are trained on subsets of the training data enhances the recognition rate.

\addtolength{\textheight}{-12cm}   
\bibliographystyle{IEEEtran}
\bibliography{refs}
\end{document}